%% file: 2020_acl_trivia_tournament.tex
\documentclass[table,11pt,a4]{article}
\usepackage[utf8]{inputenc} 
\usepackage{style/acl2020}

\newif\ifcomment\commentfalse
\input{style/preamble}

\usepackage{hyperref}
\usepackage{times}

\aclfinalcopy
\def\aclpaperid{28}

\newcommand{\figfile}[1]{2020_acl_trivia_tournament/figures/#1}
\newcommand{\autofig}[1]{2020_acl_trivia_tournament/auto_fig/#1}

\title{What Question Answering can Learn from Trivia Nerds}
\author{  \\
iSchool, \abr{cs}, \abr{umiacs}, \abr{lsc} \\
\includegraphics[scale=0.03]{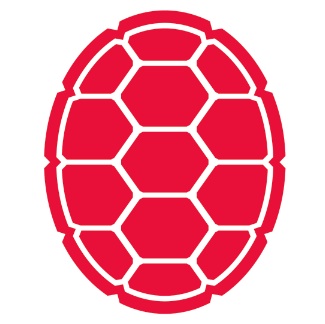} University of Maryland \\ 
\texttt{jbg@umiacs.umd.edu} \\ \And
Jordan Boyd-Graber\textsuperscript{\includegraphics[scale=0.02]{2020_acl_trivia_tournament/figures/redshell}, $\dagger$} Benjamin Börschinger$^\dagger$ \\
\\
\\
\hphantom{\dots\dots\dots\dots}\texttt{\{jbg, }
\And  
\\
\\
$\dagger$ Google Research, Z\"urich\hphantom{\dots\dots\dots\dots} \\
\texttt{bboerschinger\}@google.com}\hphantom{\dots\dots}}

\newcommand{\sota}{\abr{sota}}
\newcommand{\xlnet}{\abr{xln}{\small et}}

\begin{document}

\maketitle

\input{2020_acl_trivia_tournament/sections/00-abstract}

\input{2020_acl_trivia_tournament/sections/10-intro}
\input{2020_acl_trivia_tournament/sections/20-tournaments}
\input{2020_acl_trivia_tournament/sections/30-craft}

\input{2020_acl_trivia_tournament/sections/60-qb}

\input{2020_acl_trivia_tournament/sections/90-call}

\paragraph{Acknowledgements}

Many thanks to Massimiliano Ciaramita, Jon Clark, Christian Buck, Emily Pitler, and Michael Collins for insightful discussions that helped frame these ideas.  Thanks to Kevin Kwok for permission to use Protobowl screenshot and information.

\bibliography{bib/journal-full,bib/jbg,bib/pedro}
\bibliographystyle{style/acl_natbib}

\clearpage
\setcounter{section}{0}
\renewcommand{\thesection}{\Alph{section}}

\input{2020_acl_trivia_tournament/sections/appendix}

\end{document}

%% file: style/preamble.tex
\usepackage{framed}
\usepackage{mdwlist}
\usepackage{latexsym}
\usepackage{colortbl}
\usepackage{xcolor}
\usepackage{nicefrac}
\usepackage{booktabs}
\usepackage{amsfonts}
\usepackage[T1]{fontenc}
\usepackage{bold-extra}
\usepackage{amsmath}
\usepackage{amssymb}
\usepackage{bm}
\usepackage{graphicx}
\usepackage{mathtools}
\usepackage{microtype}
\usepackage{multirow}
\usepackage{multicol}
\usepackage{latexsym,comment}
\usepackage[normalem]{ulem}

\newcommand*{\missingreference}{{\Huge \colorbox{red}{?reference?}}}
\newcommand*{\missingcitation}{{\Huge \colorbox{red}{?citation?}}}
\makeatletter
\def\@setref#1#2#3{%
  \ifx#1\relax
    \protect\G@refundefinedtrue
    \nfss@text{\reset@font\missingreference}%
    \@latex@warning{Reference `#3' on page \thepage \space
      undefined}%
  \else
    \expandafter#2#1\null
  \fi}
\def\@citex[#1]#2{\leavevmode
  \let\@citea\@empty
  \@cite{\@for\@citeb:=#2\do
    {\@citea\def\@citea{,\penalty\@m\ }%
      \edef\@citeb{\expandafter\@firstofone\@citeb\@empty}%
      \if@filesw\immediate\write\@auxout{\string\citation{\@citeb}}\fi
      \@ifundefined{b@\@citeb}{\hbox{\reset@font\missingcitation}%
        \G@refundefinedtrue
        \@latex@warning
        {Citation `\@citeb' on page \thepage \space undefined}}%
      {\@cite@ofmt{\csname b@\@citeb\endcsname}}}}{#1}}
\makeatother

\newcommand{\gem}[1]{\mbox{\textsc{gem}}}
\newcommand{\abr}[1]{\textsc{#1}}
\newcommand{\camelabr}[2]{{\small #1}{\textsc{#2}}}

\renewenvironment{quote}
{\list{}{\rightmargin\leftmargin}%
  \item\relax\small\ignorespaces}
{\unskip\unskip\endlist}

\newcommand{\hidetext}[1]{}
\newcommand{\ignore}[1]{}

\ifcomment
  \newcommand{\pinaforecomment}[3]{\colorbox{#1}{\parbox{.8\linewidth}{#2: #3}}}
\else
  \newcommand{\pinaforecomment}[3]{}
\fi

\newcommand{\smallurl}[1]{ \begin{tiny}\url{#1}\end{tiny}}

\definecolor{lightblue}{HTML}{3cc7ea}
\definecolor{CUgold}{HTML}{CFB87C}
\definecolor{grey}{rgb}{0.95,0.95,0.95}
\definecolor{ceil}{rgb}{0.57, 0.63, 0.81}
\definecolor{UMDred}{HTML}{ed1c24}
\definecolor{UMDyellow}{HTML}{ffc20e}

\newcommand{\qb}[0]{Quizbowl}
\newcommand{\qa}[0]{\abr{qa}}
\newcommand{\triviaqa}{\camelabr{Trivia}{qa}}
\newcommand{\searchqa}{\camelabr{Search}{qa}}

\newcommand{\squad}{\textsc{sq}{\small u}\textsc{ad}}

\newcommand{\fever}{\abr{fever}}

\newcommand{\fone}{$F_1$}
\newcommand{\jeopardy}{\textit{Jeopardy!}}

%% file: 2020_acl_trivia_tournament/sections/00-abstract.tex
\begin{abstract}

In addition to machines answering
questions, question answering (\abr{qa}) research creates interesting,
challenging questions that reveal the best systems.
We argue that creating a \abr{qa} dataset---and
its ubiquitous leaderboard---closely resembles
running a trivia tournament: you write questions, have agents---humans or machines---answer questions, and declare a winner.
However, the research community has ignored the 
lessons from decades of the trivia community creating vibrant, fair,
and effective \abr{qa} competitions.
After detailing problems with existing \abr{qa} datasets, we outline several lessons that transfer to \abr{qa} research: removing ambiguity, discriminating skill,
and adjudicating disputes.

\end{abstract}

%% file: 2020_acl_trivia_tournament/sections/10-intro.tex
\section{Introduction}

This paper takes an unconventional analysis to answer ``where we've been and where we're going'' in question answering  (\abr{qa}).
Instead of approaching the question only as \abr{acl} researchers, we apply the best practices of trivia tournaments to \abr{qa} datasets.

The \qa{} community is obsessed with evaluation.
Schools, companies, and newspapers hail new \abr{sota}s and
topping leaderboards, giving rise to claims that an ``\abr{ai} model tops humans''~\citep{najberg-18} because it `won' some leaderboard, putting ``millions of jobs at risk''~\cite{cuthbertson-18}.
But what is a leaderboard? 
A leaderboard is a statistic about \qa{} accuracy that induces a ranking over participants.

Newsflash: this is the same as a trivia tournament. The trivia community has been doing this for decades~\cite{jennings-06}; 
Section~\ref{sec:tournament} details this
overlap between the qualities of a first-class \abr{qa} dataset (and
its requisite leaderboard).
The experts running these tournaments are imperfect, but they've learned from their past mistakes (see Appendix~\ref{sec:history} for a brief historical perspective) and created a community that reliably identifies those best at question answering.
Beyond the format of the \emph{competition}, important
safeguards ensure individual questions are clear, unambiguous,
and reward knowledge (Section~\ref{sec:craft}).

We are not saying that academic \abr{qa} should surrender to trivia questions or the community---far from it!
The trivia community does not understand the real world information seeking needs of users or what questions challenge computers.
However, they know how, given a bunch of questions, to declare that someone is better at answering questions than another.
This collection of tradecraft and principles can help the \abr{qa} community.

Beyond these general concepts that \abr{qa} can learn from, 
Section~\ref{sec:qb} reviews how the ``gold standard'' of trivia formats, Quizbowl, can improve traditional \abr{qa}.
We then briefly discuss how research that uses fun, fair, and good
trivia questions can benefit from the expertise, pedantry, and passion
of the trivia community (Section~\ref{sec:call}).

%% file: 2020_acl_trivia_tournament/sections/20-tournaments.tex
\section{Surprise, this is a Trivia Tournament!}
\label{sec:tournament}

``My research isn't a silly trivia tournament,'' you say.
That may be, but let us first tell you a little about what running a tournament is like, and perhaps you might see similarities.

First, the questions.
Either you write them yourself or you pay someone to write them (sometimes people on the Internet). 
There is a fixed number of questions you need to hit by a particular date. 

Then, you advertise.  
You talk about your questions:
who is writing them, what subjects are covered, and why people should try to answer them.

Next, you have the tournament. 
You keep your questions secure until test time, collect answers from all participants, and declare a winner. 
Afterward, people use the questions to train for future tournaments.

These have natural analogs to crowd sourcing questions, writing the paper, advertising, and running a leaderboard. 
The biases of academia put much more emphasis on the paper, but there are components where trivia tournament best practices could help. 
In particular, we focus on fun, well-calibrated, and discriminative tournaments.

\subsection{Are we Having Fun?}
\label{sec:fun}

Many datasets use crowdworkers to establish human accuracy~\cite{rajpurkar-16,choi-18}.
However, these are not the only humans who should answer a dataset's questions.
So should the datasets' creators.

In the trivia world, this is called a {\bf play test}:
get in the shoes of someone \emph{answering} the questions yourself.
If you find them boring, repetitive, or uninteresting, so will crowdworkers; and if you, as a human, can find shortcuts to answer questions~\cite{rondeau-18}, so will a computer.

Concretely, \newcite{weissenborn-17} catalog artifacts in \squad{}~\cite{rajpurkar-18}, arguably the most popular computer \abr{qa} leaderboard;  
and indeed, many of the questions are not particularly fun to answer from a human perspective; they're fairly formulaic.  
If you see a list like ``Along with Canada and the United Kingdom, what country\dots'', you can ignore the rest of the question and just type Ctrl+F~\cite{russell-10, yuan-19} to find the third country---\underline{Australia} in this case---that appears with ``Canada and the \abr{uk}''.
Other times, a \squad{} playtest would reveal frustrating questions that are 
i) answerable given the information in the paragraph but not with a direct span,\footnote{A source paragraph says ``In [Commonwealth countries]\dots the term is generally restricted to\dots Private education in North America covers the whole gamut\dots''; thus, the question ``What is the term private school restricted to in the US?'' has the information needed but not as a span.}
ii) answerable only given facts beyond the given paragraph,\footnote{A source paragraph says ``Sculptors [in the collection include] Nicholas Stone, Caius Gabriel Cibber, [...], \underline{Thomas Brock}, Alfred Gilbert, [...] and Eric Gill[.]'', i.e., a list of names; thus, the question ``Which British sculptor whose work includes the Queen Victoria memorial in front of Buckingham Palace is included in the V\&A collection?'' should be unanswerable in traditional machine reading.}
iii) unintentionally embedded in a discourse, resulting
in linguistically odd questions with arbitrary correct answers,\footnote{A question ``Who \emph{else} did Luther use violent rhetoric towards?'' has the gold answer ``writings condemning the Jews and in diatribes against \underline{Turks}''.}
iv)  or non-questions.

Or consider \searchqa{}~\cite{dunn-17}, derived from the game \jeopardy{}, which asks ``An article that he wrote about his riverboat days was eventually expanded into \textit{Life on the Mississippi}.''
The young apprentice and newspaper writer who wrote the article is named Samuel Clemens; however, the reference answer is that author's later pen name, \underline{Mark Twain}.
Most \qa{} evaluation metrics would count \underline{Samuel Clemens} as incorrect.
In a real game of \jeopardy{}, this would not be an issue (Section~\ref{sec:ambiguity}).

Of course, fun is relative and any dataset is bound to contain at least some errors. However, playtesting is an easy way to find systematic problems
in your dataset: unfair, unfun playtests make for ineffective leaderboards.
Eating your own dog food can help diagnose artifacts, scoring issues, or other shortcomings early in the process.

\newcite{boyd-graber-12} created an interface for play testing that was fun enough that people played for free.
After two weeks the site was taken down, but it was popular enough that the trivia community forked the open source code to create a \href{https://protobowl.com/}{bootleg version} that is still going strong almost a decade later.

The deeper issues when creating a \abr{qa} task are:
have you designed a task that is internally consistent,
supported by a scoring metric that matches your goals (more on this in a moment),
using gold annotations that correctly reward those who do the task well?
Imagine someone who loves answering the questions your task poses: would they have fun on your task?
If so, you may have a good dataset.
Gamification~\cite{ahn-06} harnesses users' passion better motivator than traditional paid labeling.
Even if you pay crowdworkers, if your questions are particularly unfun, you may need to think carefully about your dataset and your goals.

\subsection{Am I Measuring what I Care About?}
\label{subsection:measuring-what-you-care-about}

Answering questions requires multiple skills: identifying where an answer is mentioned~\cite{hermann-15}, 
knowing the canonical name for the answer~\cite{yih-15}, realizing when to answer and when to
abstain~\cite{rajpurkar-18}, or being able to justify an answer explicitly with evidence~\cite{fever-18}.
In \qa{}, the emphasis on \abr{sota} and leaderboards has focused attention on single automatically computable
metrics---systems tend to be compared by their `\squad{} score' or their `\abr{nq} score', as if this were all there
is to say about their relative capabilities.  Like \abr{qa} leaderboards, trivia tournaments need to decide
on a single winner, but they explicitly recognize that there are more interesting comparisons to be made.

For example, a tournament may recognize differnt background/resources---high school, small school, undergrads~\cite{naqt-eligibility}.  Similarly, more practical leaderboards would reflect training time
or resource requirements~\citep[see][]{dodge-19} including `constrained' or `unconstrained'
training~\citep{bojar-2014}.
Tournaments also give awards for specific skills (e.g., least incorrect
answers).  Again, there are obvious leaderboard analogs that would go beyond a single number.  For example, in \squad{} 2.0, abstaining contributes the same
to the overall \fone{} as a fully correct answer, obscuring whether a system
is more precise or an effective abstainer.  If the task
recognizes both abilities as important, reporting a single score risks implicitly prioritizing one balance of the two.

As a positive example for being explicit, the 2018 \fever{} shared task~\cite{fever-18} favors
getting the correct final answer over exhaustively justifying said answer with a  a metric that
required only one piece of evidence per question.  However, by still breaking out the full justification 
precision and recall scores in the leaderboard, it is still clear that the most precise system only was in fourth place on the ``primary'' leaderboard,
and that the top three ``primary metric'' winners are compromising on the evidence selection.

\begin{figure}[t]
    \centering
    \includegraphics[width=1.0\linewidth]{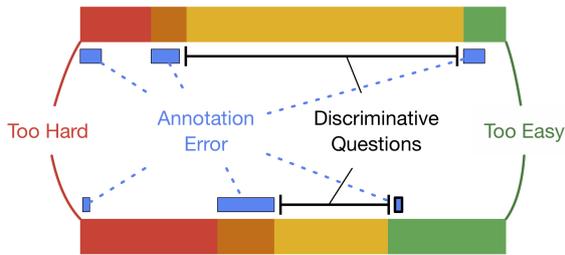}
    \caption{Two datasets with $0.16$ annotation error, but the top better discriminates \abr{qa} ability.  In the good dataset (top), most questions are challenging but not impossible.  In the bad dataset (bottom), there are more trivial or impossible questions \emph{and} annotation error is concentrated on the challenging, discriminative questions.  Thus, a smaller fraction of questions decide who sits atop the leaderboard, requiring a larger test set.}
    \label{fig:error-and-difficulty}
\end{figure}

\subsection{Do my Questions Separate the Best?}
\label{sec:discriminative}

Let us assume that you have picked a metric (or a set of metrics) that capture what you care about:
systems answer questions correctly,
abstain when they cannot,
explain why they answered the way they did,
or whatever facet of \abr{qa} is most important for your dataset.
Now, this leaderboard can rack up citations as people chase the top spot.
But your leaderboard is only useful if it is {\bf discriminative}: does it separate the best from the rest?

There are many ways questions might not be discriminative.  
If every system gets a question right (e.g., abstain on ``asdf'' or correctly answer ``What is the capital of Poland?''), it does not separate participants.  
Similarly, if every system flubs ``what is the oldest north-facing kosher restaurant'' wrong, it also does not discriminate systems.
\newcite{sugawara-18} call these questions ``easy'' and ``hard''; we instead argue for a three-way distinction.

In between easy questions (system answers correctly with probability 1.0) and hard (probability 0.0), questions with probabilities nearer to 0.5 are more interesting.
Taking a cue from Vygotsky's proximal development theory of human learning~\cite{chaiklin-03}, these discriminative questions---rather than the easy or the hard ones---should help improve \abr{qa} systems the most.
These Goldilocks\footnote{In a British folktale first recorded by Robert Southey, an interloper, ``Goldilocks'', finds three bowls of porridge: one too hot, one too cold, and one ``just right''.  Goldilocks questions' difficulty are likewise ``just right''.} questions are also most important for deciding who will sit atop the leaderboard; ideally they (and not random noise) will decide.
Unfortunately, many existing datasets seem to have many easy questions; \newcite{sugawara-20} find that ablations like shuffling word order~\cite{feng-19}, shuffling sentences, or only offering the most similar sentence do not impair systems, and \newcite{rondeau-18} hypothesize that most \abr{qa} systems do little more than pattern matching, impressive performance numbers notwithstanding.

\subsection{Why so few Goldilocks Questions?}

This is a common problem in trivia tournaments, particularly pub quizzes~\cite{diamond-09}, where too difficult questions can scare off patrons.
Many quiz masters prefer popularity over discrimination and thus prefer easier questions.

Sometimes there are fewer Goldilocks questions not by choice, but by chance: a dataset becomes less discriminative through annotation error.
All datasets have some annotation error; if this annotation error is concentrated on the Goldilocks questions, the dataset will be less useful.
As we write this in 2019, humans and computers sometimes struggle on the same questions. 
Thus, annotation error is likely to be correlated with which questions will determine who will sit atop a leaderboard.

\begin{figure*}[t!]
    \begin{center}
    \includegraphics[width=1.0\linewidth]{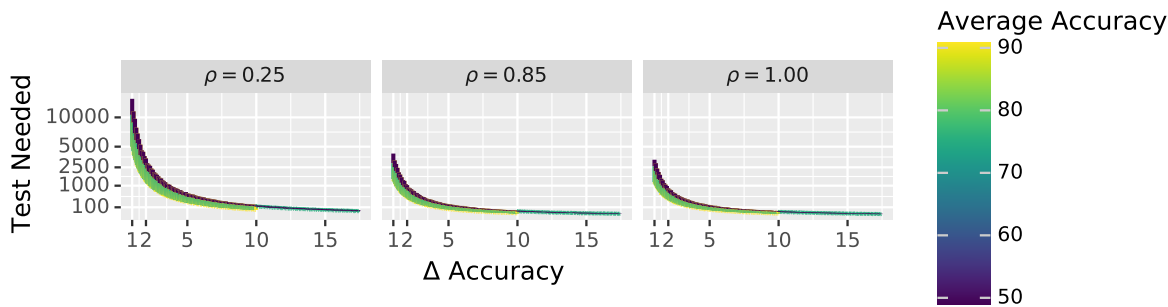}
    \end{center}
    \caption{
    How much test data do you need to discriminate two systems with 95\% confidence?  This depends on both the difference in accuracy between the systems ($x$ axis) and the average accuracy of the systems (closer to 50\% is harder).  Test set creators do not have much control over those.  They do have control, however, over how many questions are discriminative.  If all questions are discriminative (right), you only need 2500 questions, but if three quarters of your questions are too easy, too hard, or have annotation errors (left), you'll need 15000.}
    \label{fig:how-big}
\end{figure*}

Figure~\ref{fig:error-and-difficulty} shows two datsets with the same annotation error and the same number of overall questions.
However, they have different difficulty distributions and correlation of annotation error and difficulty.
The dataset that has more discriminative questions and consistent annotator error has fewer questions that are effectively useless for determining the winner of the leaderboard.
We call this the effective dataset proportion~$\rho$.
A dataset with no annotation error and only discriminative questions has $\rho=1.0$, while the bad dataset in Figure~\ref{fig:error-and-difficulty} has $\rho=0.16$.  
Figure~\ref{fig:how-big} shows the test set size required to reliably discriminate systems for different values of $\rho$, based on a simulation described in Appendix~\ref{sec:synthetic-discriminative}.

At this point, you might be despairing about how big you need your dataset to be.\footnote{Indeed, using a more sophisticated simulation approach \citet{voorhees-03} found
that the TREC 2002 QA test set could not discriminate systems with less than a seven absolute score point difference.}
The same terror transfixes trivia tournament organizers.
We discuss a technique used by them for making individual questions more discriminative using a property called pyramidality in Section~\ref{sec:pyramidality}.

%% file: 2020_acl_trivia_tournament/sections/30-craft.tex
\section{The Craft of Question Writing}
\label{sec:craft}

One thing that trivia enthusiasts agree on is that questions need to be good questions that are  well written.
Research shows that asking ``good questions'' requires sophisticated pragmatic reasoning~\cite{hawkins-15}, and pedagogy explicitly acknowledges the complexity of writing effective questions for assessing student performance~\citep[for a book length treatment of writing good multiple choice questions, see]{Haladyna-04}.  There is no question that writing good questions is a craft in its own right.

\abr{qa} datasets, however, are often collected from the wild or written by untrained
crowdworkers.
Even assuming they are confident users of English (the primary language
of \abr{qa} datasets), crowdworkers lack experience in 
crafting questions and may introduce idiosyncracies that shortcut machine learning~\citep{geva-19}.  
Similarly, data collected from the wild
such as Natural Questions~\citep{kwiatkowski-19} by design have vast variations in quality.
In the previous section, we focused on how datasets as a whole should be structured.
Now, we focus on how specific \emph{questions} should be structured to make the dataset as valuable as possible.

\subsection{Avoiding Ambiguity and Assumptions}
\label{sec:ambiguity}

Ambiguity in questions not only frustrates answerers who resolved the ambiguity `incorrectly`, they generally undermine the usefulness of questions to precisely assess knowledge at all.  For this reason, the \abr{us} Department of Transportation explicitly bans ambiguous questions from exams for flight instructors~\cite{dot-08}; and the trivia community has developed rules that prevent ambiguity from arising.
While this is true in many contexts, examples are rife in format called \qb{}~\cite{boyd-graber-12}, whose very long questions\footnote{Like \jeopardy{}, they are not syntactically questions but still are designed to elicit knowledge-based responses; for consistency, we will still call them questions.} showcase trivia writers' tactics.
For example, Zhu Ying (2005 \abr{parfait}) warns [emphasis added]:
\begin{quote}
 He's {\bf not Sherlock Holmes}, but his address is 221B. He's {\bf not the Janitor on Scrubs}, but his father is played by R. Lee Ermy. [\dots] For ten points, name this misanthropic, crippled, Vicodin-dependent central character of a FOX medical drama. \\
{\bf ANSWER:} Gregory \underline{House}, MD
\end{quote}
to head off these wrong answers.  

In contrast, \qa{} datasets often contain ambiguous and under-specified questions.  While this sometimes
reflects real world complexities such as actual under-specified or ill-formed search queries~\citep{faruqui-18,kwiatkowski-19}, simply ignoring this ambiguity is problematic. 
As a concrete example, 
consider Natural Questions~\cite{kwiatkowski-19} where the gold answer to ``what year did the us hockey team won [sic] the olympics'' has the answers \underline{1960} and \underline{1980}, ignoring the \abr{us} women's team, which won in 1998 and 2018, and further assuming the query is about ice rather than field hockey (also an olympic sport).
This ambiguity is a \textit{post hoc} interpretation as the associated Wikipedia pages are, indeed, about the 
United States men's national ice hockey team.
We contend the ambiguity persists in the original question, and the information retrieval arbitrarily provides one of many interpretations.
True to its name, these under-specified queries make up a substantial 
(if not almost the entire) fraction of natural questions in search contexts.

The problem is neither that such questions exist nor that \abr{mrqa} considers
questions relative to the associated context. 
The problem is that tasks do not explicitly
acknowledge the original ambiguity and instead gloss over the implicit assumptions used in creating the
data.
This introduces potential noise and bias (i.e., giving a bonus to systems that make
the same assumptions as the dataset) in leaderboard rankings. 
At best, these will become part of
the measurement error of datasets (no dataset is perfect). 
At worst, they will recapitulate the biases that went into the creation of the datasets.
Then, the community will implicitly equate the biases with correctness: you get high scores if you 
adopt this set of assumptions.
Then, these enter into real-world systems, further perpetuating the bias.
Playtesting can reveal these issues (Section~\ref{sec:fun}), as implicit assumptions 
can rob a player of correctly answered questions.
If you wanted to answer \underline{2014} to ``when did Michigan last win the championship''---when the Michigan State Spartans won the Women's Cross Country championship---and you cannot because you chose the wrong school, the wrong sport, and the wrong gender,
you would complain as a player; researchers instead discover latent assumptions that crept into the data.\footnote{Where to draw the line is a matter of judgment; computers---who lack common sense---might find questions ambiguous where humans would not.}

It is worth emphasizing that this is not a purely hypothetical problem. For example, Open Domain Retrieval Question Answering~\cite{lee-19} deliberately avoids providing a reference context for the question in its framing but, in re-purposing data such as Natural Questions, opaquely relies on it for the gold answers.

\subsection{Avoiding superficial evaluations}

A related issue is that, in the words of \citet{voorhees-00}, ``there is no such
thing as a question with an obvious answer''.
As a consequence, trivia question authors are considered to
delineate acceptable and unacceptable answers.

For example, Robert Chu (Harvard Fall XI) uses a mental model of an answerer to explicitly delineate the range of acceptable correct answers:
\begin{quote}
     In Newtonian gravity, this quantity satisfies Poisson's equation. [\dots] For a dipole, this quantity is given by negative the dipole moment dotted with the electric field. [\dots] For 10 points, name this form of energy contrasted with kinetic.\\
    {\bf ANSWER:} \underline{potential energy} \textit{(prompt on energy; accept specific types like electrical potential energy or gravitational potential energy; do not accept or prompt on just ``potential'')}
\end{quote}

Likewise, the style guides for writing questions stipulate that you
must give the answer type clearly and early on.
These mentions specify whether you want a book, a collection, a movement, etc.  
It also signals the level of specificity requested.  
For example, a question about a date must state ``day and month required'' (\underline{September 11}, ``month and year required'' (\underline{April 1968}), or ``day, month, and year required'' (\underline{September 1, 1939}).
This is true for other answers as well: city and team, party and country, or more generally ``two answers required''.
Despite all of these conventions, no pre-defined set of answers is perfect, 
and a process for adjudicating answers is an integral part of trivia competitions.

In major high school and college national competitions and game shows, if low-level staff cannot resolve the issue by either throwing out a single question or accepting minor variations (\underline{America} instead
of \underline{\abr{usa}}), the low-level staff contacts the tournament director.
The tournament director---with their deeper knowledge of rules and questions---often decide the issue.
If not, the protest goes
through an adjudication process designed to minimize
bias:\footnote{\url{https://www.naqt.com/rules/\#protest}}
write the summary of the dispute,
get all parties to agree to the summary,
and then hand the decision off to mutually agreed experts from the tournament's phone tree.
The substance of the disagreement is communicated (without identities), and the experts apply the rules and decide.

For example, a 
when a particularly inept \jeopardy{} contestant\footnote{\url{http://www.j-archive.com/showgame.php?game_id=6112}} answered \underline{endoscope} to ``Your surgeon could choose to take a look inside you with this type of fiber-optic instrument''.
Since the van Doren scandal~\cite{freedman-97}, every television trivia contestant has an advocate assigned from an
auditing company.
In this case, the advocate initiated a process that went to a panel of judges who then
ruled that \underline{endoscope} (a more general term) was also correct.

The need for a similar process seems to have been well-recognized
in the earliest days of \qa{} system bake-offs such as \abr{trec-qa}, and \newcite{voorhees-08} notes that
\begin{quote}
    [d]ifferent \abr{qa} runs very seldom return exactly the same [answer], and it is quite difficult to determine automatically whether the difference [\dots] is significant.
\end{quote} 
In stark contrast to this, \qa{} datasets typically only provide a single string or, if one is lucky,
several strings.  
A correct answer means \emph{exactly} matching these strings or at least
having a high token overlap \fone{}, and failure to agree with the
pre-recorded admissible answers will put you at an uncontestable
disadvantage on the leaderboard (Section~\ref{subsection:measuring-what-you-care-about}).  

To illustrate how current evaluations fall short of meaningful discrimination,
we qualitatively analyze two near-\sota{} systems on
\squad{} V1.1 on CodaLab:
the original \xlnet{}~\citep{yang-19} and a subsequent iteration
called \xlnet{}-123.\footnote{We could not find a paper describing \xlnet{}-123 in detail, the submission is by  \url{http://tia.today}.} 

Despite \xlnet{}-123's margin of almost four absolute \fone{} 
($94$ vs $98$) on development data, a manual inspection of a sample
of 100 of \xlnet{}-123's wins indicate that around two-thirds are `spurious':  $56\%$ are likely to be considered not only equally
good but essentially identical;  $7\%$ are cases where the answer set ommits a correct alternative;  and $5\%$ of cases are `bad' questions.\footnote{Examples in Appendix~\ref{sec:qualitative-analysis}.}

Our goal is not to dwell on the exact proportions, to minimize the achievements of these strong systems, or to minimize the usefulness of quantitative evaluations.
We merely want to raise the limitation of `blind automation' for distinguishing between
systems on a leaderboard. 

Taking our cue from the trivia community, here is an alternative
for \abr{mrqa}.
Test sets are only created for a specific time; all systems are submitted simultaneously.
Then, all questions and answers are revealed.
System authors can protest correctness rulings on questions, directly
addressing the issues above. 
After agreement is reached,
quantitative metrics are computed for comparison purposes---despite their inherent limitations they at least can be
trusted.
Adopting this for \abr{mrqa} would require creating a new, smaller test
set every year.
However, this would gradually refine the annotations and process.

This suggestion is not novel: \citet{voorhees-00} conclude that automatic evaluations
are sufficient ``for experiments internal to an organization where the
benefits of a reusable test collection are most significant (\emph{and
the limitations are likely to be understood})''~(our emphasis) but
that ``satisfactory techniques for [automatically] evaluating new
runs'' have not been found yet.  We are not aware of any change on
this front---if anything, we seem to have become more insensitive
as a community to just how limited our current evaluations are.

\subsection{Focus on the Bubble}

While every question should be perfect, time and resources are limited.  
Thus, authors of tournaments have a policy of ``focusing on the bubble'', where the ``bubble'' are the questions mostly likely to discriminate between top teams.

For humans, authors and editors focus on the questions and clues that they predict will decide the tournament.
These questions are thoroughly playtested, vetted, and edited.
Only after these questions have been perfected will the other questions undergo the same level of polish.

For computers, the same logic applies.  
Authors should ensure that these questions are correct, free of ambiguity, and unimpeachable.
However, as far as we can tell, the authors of \qa{} datasets do not give any special attention to these questions.

Unlike a human trivia tournament, however---with finite patience of the participants---this does not mean that you should necessarily remove all of the easy or hard questions from your dataset---spend more of your time/effort/resources on the bubble.
You would not want to introduce a sampling bias that leads to inadvertently forgetting how to answer questions like ``who is buried in Grant's tomb?''~\cite[Chapter 7]{dwan-00}.

%% file: 2020_acl_trivia_tournament/sections/60-qb.tex
\section{Why \qb{} is the Gold Standard}
\label{sec:qb}

We now focus our thus far wide-ranging \abr{qa} discussion to a specific format: \qb{}, which has many of the desirable properties outlined above.
We have no delusion that mainstream \abr{qa} will universally adopt this format.
However, given then community's emphasis on fair evaluation, computer \abr{qa} can borrow \emph{aspects} from the gold standard of human \abr{qa}.
We discuss what this may look like in Section~\ref{sec:call}, but first we describe the gold standard of human \abr{qa}.

We have shown several examples of \qb{} questions, but we have not yet explained in detail how the format works; see \newcite{DBLP:journals/corr/abs-1904-04792} for a more comprehensive description.
You might be scared off by how long the questions are.
However, in real \qb{} trivia tournaments, they are not finished because the questions are designed to be \emph{interrupted}.

\paragraph{Interruptable}

A moderator reads a question.
Once someone knows the answer, they use a signaling device to ``\emph{buzz in}''.
If the player who buzzed is right, they get points.
Otherwise, they lose points and the question continues for the other team.  

Not all trivia games with buzzers have this property, however.
For example, take \jeopardy{}, the subject of Watson's \textit{tour de force}~\cite{ferruci-10}.  
While \jeopardy{} also uses signaling devices, these only work \emph{at the end of the question}; Ken Jennings, one of the top \jeopardy{} players (and also a \qb{}er) explains it on a \textit{Planet Money} interview~\cite{malone-19}:
\begin{quote}
{\bf Jennings:} The buzzer is
    not live until Alex finishes reading the question. And if you buzz
    in before your buzzer goes live, \emph{you actually lock yourself out
    for a fraction of a second}. So the big mistake on the show is
    people who are all adrenalized and are buzzing too quickly, too
    eagerly. \\
{\bf Malone:} OK. To some degree, \jeopardy{} is kind of a video game, and a \emph{crappy video game where it's, like, light goes on, press button}---that's it. \\
{\bf Jennings:} (Laughter) Yeah. \\
\end{quote}
Thus, \jeopardy{}'s buzzers are a gimmick to ensure good television; however, \qb{} buzzers discriminate knowledge (Section~\ref{sec:discriminative}).
Similarly, while \triviaqa{}~\cite{joshi-17} is written by knowledgeable writers, the questions are not pyramidal.

\paragraph{Pyramidal}
\label{sec:pyramidality}

Recall that an effective dataset (tournament) discriminates the best from the rest---the higher the proportion of effective questions~$\rho$, the better.
\qb{}'s $\rho$ is nearly 1.0 because discrimination happens \emph{within} a question: after every word, an answerer must decide whether they have enough information to answer the question.
\qb{} questions are arranged so that questions are maximally \emph{pyramidal}: questions begin with hard clues---ones that require deep understanding---to more accessible clues that are well known.

\paragraph{Well-Edited}

\qb{} questions are created in phases.
First, the \emph{author} selects the answer and assembles (pyramidal) clues.
A \emph{subject editor} then removes ambiguity, adjusts acceptable answers, and tweaks clues to optimize discrimination.
Finally, a \emph{packetizer} ensures the overall set is diverse, has uniform difficulty, and is without repeats.

\paragraph{Unnatural}
\label{sec:unnatural}

Trivia questions are fake: the asker already knows the answer.  
But they're no more fake than a course's final exam, which like leaderboards are designed to test knowledge.

Experts know when questions are ambigiuous; while ``what play has a character whose father is dead'' could be \textit{Hamlet}, \textit{Antigone}, or \textit{Proof}, a good writer's knowledge avoids the ambiguity.
When authors omit these cues, the question is derided as a ``hose''~\cite{2013-eltinge}, which robs the tournament of fun (Section~\ref{sec:fun}).

One of the benefits of contrived formats is a focus on specific phenomena. 
\newcite{dua-19} exclude questions an existing \abr{mrqa} system could answer to focus on challenging quantitative reasoning.
One of the trivia experts consulted in \newcite{wallace-19} crafted a question that tripped up neural \abr{qa} systems with ``this author opens Crime and Punishment''; the top system confidently answers \underline{Fyodor Dostoyevski}.
However, that phrase was embedded in a longer question ``The narrator in \textit{Cogwheels} by this author opens \textit{Crime and Punishment} to find it has become \textit{The Brothers Karamazov}''. 
Again, this shows the inventiveness and linguistic dexterity of the trivia community.

A counterargument is that when real humans ask questions---e.g., on Yahoo! Questions~\cite{szpektor-13},
Quora~\cite{iyer-17} or web search~\cite{kwiatkowski-19}---they ignore the craft of question writing.
Real humans tend to react to unclear questions with confusion or divergent answers, often making
explicit in their answers how they interpreted the original question (``I assume you  meant\dots'').

Given real world applications will have to deal with the inherent noise and ambiguity of unclear questions, our systems must cope with it.  However, dealing with it
is not the same as glossing over their complexities.

\paragraph{Complicated} 

\qb{} is more complex than other datasets.  
Unlike other datasets where you just need to decide \emph{what} to answer, you also need to choose \emph{when} to answer the question.
While this improves the dataset's discrimination, it can hurt popularity because you cannot copy/paste code from other \abr{qa} tasks.
The complicated pyramidal structure also makes some questions---e.g., what is log base four of sixty-four---difficult\footnote{But not always impossible, as \abr{ihssbca} shows:
\begin{quote}
    This is the smallest counting number which is the radius of a sphere whose volume is an integer multiple of $\pi$. It is also the number of distinct real solutions to the equation $x^7-19x^5=0$. This number also gives the ratio between the volumes of a cylinder and a cone with the same heights and radii. Give this number equal to the log base four of sixty-four.
\end{quote}} to ask.
However, the underlying mechanisms (e.g., reinforcement learning) share properties with other tasks, such as simultaneous translation~\cite{grissom:he:boyd-graber:morgan-2014,ma-etal-2019-stacl}, human incremental processing~\cite{levy-08,levy-11}, and opponent modeling~\cite{he-16}.


%% file: 2020_acl_trivia_tournament/sections/90-call.tex
\section{A Call to Action}
\label{sec:call}

You may disagree with the superiority of \qb{} as a \qa{} framework (even for trivia nerds, not all agree\dots \textit{de gustibus non est disputandum}).
In this final section, we hope to distill our advice into a call to action regardless of your question format of choice.
Here are our recommendations if you want to have an effective leaderboard.

\paragraph{Talk to Trivia Nerds}

You should talk to trivia nerds because they have useful information (not just about the election of 1876).
Trivia is not just the accumulation of information but also connecting disparate facts~\cite{jennings-06}.
These skills are exactly those that we want computers to develop.

Trivia nerds are writing questions anyway; we can save money and time if we pool resources.
Computer scientists benefit if the trivia community writes questions that aren't trivial for computers to solve (e.g., avoiding quotes and named entities).
The trivia community benefits from tools that make their job easier: show related questions, link to Wikipedia, or predict where humans will answer.

Likewise, the broader public has unique knowledge and skills.
In contrast to low-paid crowdworkers, public platforms for question answering and citizen science~\cite{bowser-13} are brimming with free expertise if you can engage the relevant communities.
For example, the \text{Quora} query ``Is there a nuclear control room on nuclear aircraft carriers?'' is purportedly answered by someone who worked in such a room~\cite{humphries-17}.
As machine learning algorithms improve, the ``good enough'' crowdsourcing that got us this far may simply not be enough for continued progress.

Many question answering datasets benefit from the efforts of the trivia community.
Ethically using the data requires acknowledging their contributions and using their input to create datasets~\cite[Consent and Inclusivity]{jo-20}.

\paragraph{Eat Your Own Dog Food}

As you develop new question answering tasks, you should feel comfortable playing the task as a human.
Importantly, this is not just to replicate what crowdworkers are doing (also important) but to remove hidden assumptions, institute fair metrics, and define the task well.
For this to feel real, you will need to keep score; have all of your coauthors participate and compare their scores.

Again, we emphasize that {\bf human and computer skills are not identical}, but this is a benefit: humans natural aversion to unfairness will help you create a better task, while computers will blindly optimize a broken objective function~\cite{bostrom-03,Manheim-18}.
As you go through the process of playing on your question--answer dataset, you can see where you might have fallen short on the goals we outline in Section~\ref{sec:craft}.

\paragraph{Won't Somebody Look at the Data?}

After \abr{qa} datasets are released, there should also be deeper, more frequent discussion of actual questions within the \abr{nlp} community.
Part of every post-mortem of trivia tournaments is a detailed discussion of the questions, where good questions are praised and bad questions are excoriated.
This is not meant to shame the writers but rather to help build and reinforce cultural norms: questions should be well-written, precise, and fulfill the creator's goals.
Just like trivia tournaments, \abr{qa} datasets resemble a product for sale.
Creators want people to invest time and sometimes money (e.g., \abr{gpu} hours) in using their data and submitting to their leaderboards.
It is ``good business'' to build a reputation for quality questions and discussing individual questions.

Similarly, discussing and comparing the actual predictions made by the competing systems should be part of
any competition culture---without it, it is hard to tell what a couple of points
on some leaderboard mean.  To make this possible, we recommend that leaderboards include an
easy way for anyone to download a system's development predictions for qualitative analyses.

\paragraph{Make Questions Discriminative}

We argue that questions should be discriminative (Section~\ref{sec:discriminative}), and while \qb{} is one solution (Section~\ref{sec:qb}), not everyone is crazy enough to adopt this (beautiful) format.
For more traditional \abr{qa} tasks, you can maximize the usefulness of your dataset by ensuring as many questions as possible are challenging (but not impossible) for today's \abr{qa} systems.

But you can use some \qb{} intuitions to improve discrimination.
In visual \abr{qa}, you can offer increasing resolutions of the image.
For other settings, create pyramidality by adding metadata: coreference, disambiguation, or alignment to a knowledge base.
In short, consider multiple versions/views of your data that progress from difficult to easy.
This not only makes more of your dataset discriminative but also reveals what makes a question answerable.

\paragraph{Embrace Multiple Answers or Specify Specificity}

As \qa{} moves to more complicated formats and answer candidates, what constitutes a correct answer becomes more complicated.
Fully automatic evaluations are valuable for both training and quick-turnaround evaluation.
In the case annotators disagree, the question should explicitly state what level of specificity is required (e.g., \underline{September 1, 1939} vs. \underline{1939} or \underline{Leninism} vs. \underline{socialism}).
Or, if not all questions have a single answer, link answers to a knowledge base with multiple surface forms or explicitly enumerate which answers are acceptable.

\paragraph{Appreciate Ambiguity}

If your intended \abr{qa} application has to handle ambiguous questions,
do justice to the ambiguity by making it part of your task---for example, recognize the
original ambiguity and resolve it (``did you mean\dots'') instead of giving credit
for happening to `fit the data'.

To ensure that our datasets properly ``isolate the property that motivated it in the first place''~\cite{Zaenen-2006}, we need to explicitly appreciate the unavoidable ambiguity instead of silently glossing over it.\footnote{Not surprisingly, `inherent' ambiguity is not limited to \abr{qa}; \citet{pavlick-19} show natural language inference data have `inherent disagreements' between humans and advocate for recovering the full range of accepted inferences.}

This is already an active area of research, with conversational \abr{qa} being a new setting
actively explored by several datasets~\cite{reddy-18,choi-18}; and other work explicitly focusing on
identifying useful clarification questions~\cite{rao-2018}, thematically linked questions~\cite{elgohary-18} or resolving ambiguities that arise from
coreference or pragmatic constraints by rewriting underspecified question strings in context~\cite{elgohary-19}.

\paragraph{Revel in Spectacle}

However, with more complicated systems and evaluations, a return to the yearly evaluations of \abr{trecqa} may be the best option.
This improves not only the quality of evaluation (we can have real-time human judging) but also lets the test set reflect the build it/break it cycle~\cite{ruef-16}, as attempted by the 2019 iteration of \abr{fever}.
Moreover, another lesson the \abr{qa} community could learn from trivia games is to turn it into a spectacle: exciting games with a telegenic host.
This has a benefit to the public, who see how \abr{qa} systems fail on difficult questions and to \abr{qa} researchers, who have a spoonful of fun sugar to inspect their systems' output and their competitors'.

In between are automatic metrics that mimic the flexibility of human raters, inspired by machine translation evaluations~\cite{papineni-02,specia-10} or summarization~\cite{lin-04}.  However, we should not forget
that these metrics were introduced as `understudies'---good enough when quick evaluations are needed for system
building but no substitute for a proper evaluation.
In machine translation, \newcite{laubli-20} reveal that crowdworkers cannot spot the errors that neural \abr{mt} systems make---fortunately, trivia nerds are cheaper than professional translators.

\paragraph{Be Honest in Crowning \abr{qa} Champions}

While---particularly for leaderboards---it is tempting to turn everything into a single number, recognize that there are often different sub-tasks and types of players who deserve recognition.
A simple model that requires less training data or runs in under ten milliseconds may be objectively more useful than a bloated, brittle monster of a system that has a slightly higher \fone{}~\cite{dodge-19}. 
While you may only rank by a single metric (this is what trivia tournaments do too), you may want to recognize the highest-scoring model that was built by undergrads, took no more than one second per example, was trained only on Wikipedia, etc.

Finally, if you want to make human--computer comparisons, pick the right humans.  Paraphrasing
a participant of the 2019 \abr{mrqa} workshop~\cite{fisch-19}, a system better than the average human at brain surgery does not imply superhuman performance in brain surgery.  
Likewise, beating a distracted crowdworker on \abr{qa} is not \abr{qa}'s endgame.  
If your task is realistic, fun, and challenging, you will find experts to play against your computer.
Not only will this give you human baselines worth reporting---they can also tell you how to fix your \abr{qa} dataset\dots after all, they've been at it longer than you have.

%% file: 2020_acl_trivia_tournament/sections/appendix.tex
\section*{Appendix}

Figure, footnote, and table numbers continue from main article.

\section{An Abridged History of Modern Trivia}
\label{sec:history}

In the United States, modern trivia exploded immediately after World War II via countless game shows including College Bowl~\cite{Baber-15}, the precursor to \qb{}.
The craze spread to the United Kingdom in a bootlegged version of \qb{} called \textit{University Challenge} (now liscensed by \abr{itv}) and pub quizzes~\cite{taylor_mcnulty_meek}.

The initial explosion, however, was not without controversy.
A string of cheating scandals, most notably the Van Doren~\cite{freedman-97} scandal (the subject of the film \textit{Quiz Show}), and the 1977 entry of \qb{} into intercollegiate competition forced trivia to ``grow up''.
Professional organizations and more ``grownup'' game shows like \jeopardy{} (the ``all responses in the form of a question'' gimmick grew out of how some game shows gave contestants the answers) helped created formalized structures for trivia.

As the generation that grew up with formalized trivia reached adulthood, they sought to make the outcomes of trivia competitions more rigorous, eschewing the randomness that makes for good television.
Organizations like National Academic Quiz Tournaments and the Academic Competition Federation created routes for the best players to help direct how trivia competitions would be run.
In 2019, these organizations have institutionalized the best practices of ``good trivia'' described here.

\section{Simulating the Test Set Needed}
\label{sec:synthetic-discriminative}

We simulate a head-to-head trivia competition where System~A and System~B have an accuracy $a$ (probability of getting a question right) separated by some difference: $a_A - a_B \equiv \Delta$.
We then simulate this on a test set of size $N$---scaled by the effective dataset proportion $\rho$---via draws from two Binomial distributions with success probabilities of $a_{A}$ and $a_{B}$, respectively:  
\begin{align}
    R_a \sim & \mbox{Binomial}(\rho N, a_A \notag \\
    R_b \sim & \mbox{Binomial}(\rho N, a_B)
    \label{eq:two-systems}
\end{align}
and see the minimum test set questions (using an experiment size of 5000) needed to detect the better system 95\% of the time (i.e., the minimum $N$ such that $R_a > R_b$ from Equation~\ref{eq:two-systems} in $0.95$ of the experiments).
Our emphasis, however is $\rho$: the smaller the percentage of discriminative questions (either because of difficulty or because of annotation error), the larger your test set must be.\footnote{Disclaimer: This should be only one of many considerations in deciding on the size of your test set.  Other factors may include balancing for demographic properties, covering linguistic variation, or capturing task-specific phenomena.}

\section{Qualitative Analysis Examples}
\label{sec:qualitative-analysis}

We provide some concrete examples for the classes into which we classified the
\xlnet{}-123 wins over \xlnet{}.  We indicate \underline{gold answer spans} (provided by the
human annotators) by underlining (there may be, \textbf{the \xlnet{} answer span} by bold face,
and \textit{the \xlnet{}-123 answer span} by italics, \textbf{combining} \underline{\textit{\textbf{for} tokens shared}} \textit{between spans} as is appropriate.

\subsection{Insignificant and significant span differences}
\begin{quote}
    {\bf QUESTION:}
    What type of vote must the Parliament have to either block or suggest changes to the Commission's proposals? \\
    {\bf CONTEXT:} 
    [\dots] The essence is there are three readings, starting with \underline{a Commission proposal}, where the Parliament must vote by \textbf{\underline{a \textit{majority}} of all MEPs} (not just those present) to block or suggest changes, [\dots]
\end{quote}
\textbf{\underline{a majority} of all MEPs} is as good an answer as \textit{\underline{majority}},
yet its Exact Match score is $0$.  Note that the problem is not merely one of picking a soft metric;
even its Token-F1 score is merely  $0.4$, effectively penalizing a system for giving a more complete
answer.  The limitations of Token-F1 become even clearer in light of the following significant span difference:
\begin{quote}
    {\bf QUESTION:}
    What measure of a computational problem broadly defines the inherent difficulty of the solution?
 \\
    {\bf CONTEXT:} 
    A problem is regarded as inherently difficult \underline{\textit{if its solution requires \textbf{significant resources}}}, whatever the algorithm used. [\dots]
\end{quote}

We agree with the automatic eval that a system answering \underline{\textbf{significant resources}} to this question
should not be given full (and possibly no) credit as it fails to mention relevant context.  Nevertheless,
the Token-F1 of this answer is $0.57$, i.e. larger than for the insignificant span difference just
discussed.

\subsection{Missing Gold Answers}

We also observed $7$ (out of $100$) cases of missing gold answers.  As an example, consider
\begin{quote}
    {\bf QUESTION:}
    What would someone who is civilly disobedient do in court?
 \\
    {\bf CONTEXT:} 
    Steven Barkan writes that if defendants \textit{\underline{plead not guilty}}, ``they must decide whether their primary goal will be to win an acquittal and avoid imprisonment or a fine, or to use the proceedings as a forum to \uline{inform the jury and the public of the political circumstances} surrounding the case and their reasons for breaking the law via civil disobedience.'' [\dots] \\
    In countries such as the United States whose laws guarantee the right to a jury trial but do not excuse lawbreaking for political purposes, some civil disobedients seek \textbf{jury nullification}.
\end{quote}

While annotators did mark two distinct spans as gold answers, they happen to have not marked up \textbf{jury nullification}
which is a fine answer to the question and should be rewarded.  Any individual case can likely be discussed as to
whether this truly is a missing answer or the question rules it out due to some specific subtlety in its phrasing.
This is precisely the point---relying on a pre-collected set of static answer strings without a process for
adjudicating cases of disagreements in official comparisons does not do justice to the complexity of question answering.

\subsection{Bad Questions}

We also observed $5$ cases of genuinely bad questions.  Consider
\begin{quote}
    {\bf QUESTION:}
    What library contains the Selmur Productions catalogue?
 \\
    {\bf CONTEXT:} 
    Also part of \textbf{the library} is the aforementioned \underline{\textit{Selznick}} library, the Cinerama Productions/Palomar theatrical library and the Selmur Productions catalog that the network acquired some years back [\dots]
\end{quote}

This is simply an annotation error---the correct answer to the question is not available
from the paragraph and would have to be (the American Broadcast Company's) Programming Library.
While we have to live with annotation errors as part of reality, it is not clear that we 
ought to simply accept them for \emph{official evaluations}---any human taking a closer look
at the paragraph, as part of an adjudication process, would at least note that the question is
problematic.

Other cases of `annotation' error are more subtle, involving meaning-changing typos, for example:

\begin{quote}
    {\bf QUESTION:}
    Which French kind [sic] issued this declaration?
 \\
    {\bf CONTEXT:} 
    They retained the religious provisions of the Edict of Nantes until the rule of \textit{\uline{Louis XIV}}, who progressively increased persecution  of them until he issued the Edict of Fontainebleau (1685), which abolished all legal recognition of \textbf{Protestantism} in France[.] [...]
\end{quote}

While one could debate whether or not systems ought to be able to do
`charitable' reinterpretations of the question text, this is part of the point---cases like these warrant discussion and should not be silently glossed over when `computing the score'.